\documentclass[11pt,a4paper]{article}
\usepackage[hyperref]{acl2020}
\usepackage{times}
\usepackage{latexsym}

\usepackage{microtype}
\usepackage{amssymb}
\usepackage{amsmath}
\usepackage{blindtext}
\usepackage{booktabs}
\usepackage{multirow}
\usepackage{graphicx}
\usepackage{subcaption}
\usepackage{algorithm}
\usepackage{algorithmic}
\usepackage{enumitem}

\aclfinalcopy 
\def\aclpaperid{349} 

\newcommand\mcal[1]{\mathcal{#1}}
\newcommand\name{DeeBERT}
\newcommand\highway{off-ramp}
\newcommand\highways{off-ramps}

\title{\name: Dynamic Early Exiting for Accelerating BERT Inference}

\author{Ji Xin$^{1,2}$, Raphael Tang$^{1,2}$, Jaejun Lee$^{1}$, Yaoliang Yu$^{1,2}$, \and Jimmy Lin$^{1,2}$ \vspace{5pt} \\
	$^1$David R. Cheriton School of Computer Science, University of Waterloo \\
	$^2$Vector Institute for Artificial Intelligence  \vspace{5pt}  \\ 
	{\tt   \{ji.xin,r33tang,j474lee,yaoliang.yu,jimmylin\}@uwaterloo.ca}
}

\date{}

\begin{document}
\maketitle

\begin{abstract}
Large-scale pre-trained language models such as BERT have brought significant improvements to NLP applications.
However, they are also notorious for being slow in inference, which makes them difficult to deploy in real-time applications.
We propose a simple but effective method, \name, to accelerate BERT inference.
Our approach allows samples to exit earlier without passing through the entire model.
Experiments show that \name~is able to save up to $\sim$40\% inference time with minimal degradation in model quality.
Further analyses show different behaviors in the BERT transformer layers and also reveal their redundancy.
Our work provides new ideas to efficiently apply deep transformer-based models to downstream tasks.
Code is available at \url{https://github.com/castorini/DeeBERT}.
\end{abstract}


\section{Introduction}

Large-scale pre-trained language models such as ELMo~\cite{peters-etal-2018-deep}, GPT~\cite{radford2019language}, BERT~\cite{devlin-etal-2019-bert}, XLNet~\cite{Yang2019XLNetGA}, and RoBERTa~\cite{liu2019roberta} have brought significant improvements to natural language processing (NLP) applications.
Despite their power, they are notorious for being enormous in size and slow in both training and inference.
Their long inference latencies present challenges to deployment in real-time applications and hardware-constrained edge devices such as mobile phones and smart watches.

To accelerate inference for BERT, we propose \textbf{\name}:\ \textbf{D}ynamic \textbf{e}arly \textbf{e}xiting for \textbf{BERT}.
The inspiration comes from a well-known observation in the computer vision community:\ in deep convolutional neural networks, higher layers typically produce more detailed and finer-grained features~\cite{zeiler2014visualizing}.
Therefore, we hypothesize that, for BERT, features provided by the intermediate transformer layers may suffice to classify some input samples.

\begin{figure}[t]
    \centering
    \includegraphics[width=\columnwidth]{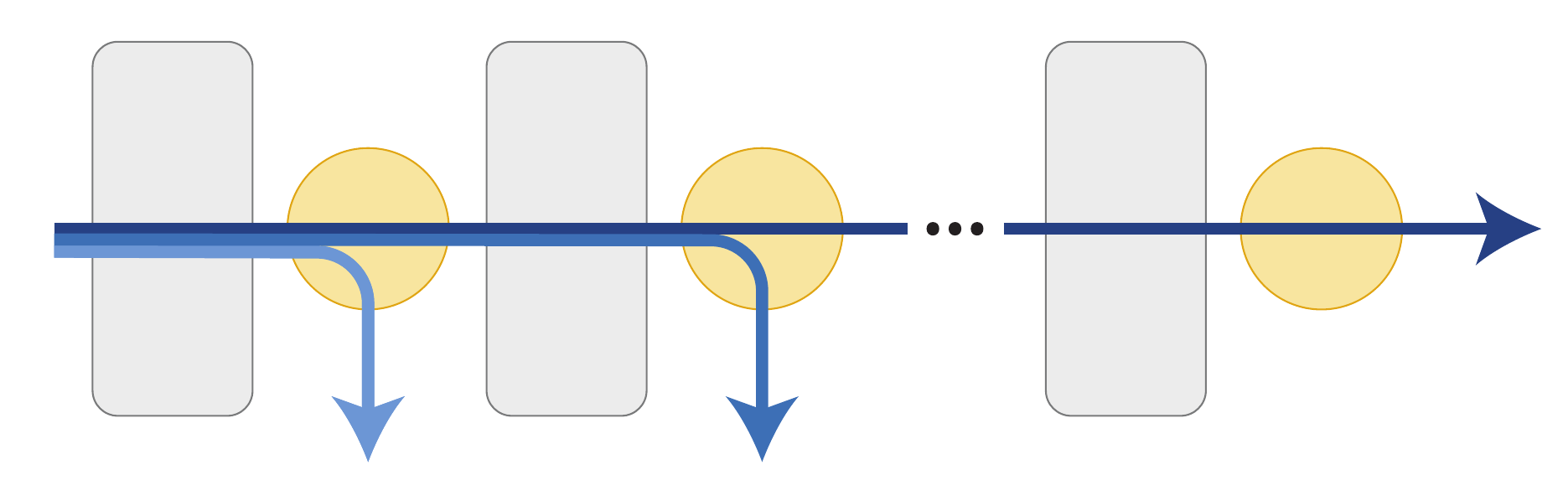}
    \caption{\name~model overview. Grey blocks are transformer layers, orange circles are classification layers (\highways), and blue arrows represent inference samples exiting at different layers.}
    \label{fig:model}
\end{figure}

\name~accelerates BERT inference by inserting extra classification layers (which we refer to as \textit{\highways}) between each transformer layer of BERT (Figure \ref{fig:model}).
All transformer layers and \highways~are jointly fine-tuned on a given downstream dataset.
At inference time, after a sample goes through a transformer layer, it is passed to the following \highway.
If the \highway~is confident of the prediction, the result is returned; otherwise, the sample is sent to the next transformer layer.

In this paper, we conduct experiments on BERT and RoBERTa with six GLUE datasets, showing that \name~is capable of accelerating model inference by up to $\sim$40\% with minimal model quality degradation on downstream tasks.
Further analyses reveal interesting patterns in the models' transformer layers, as well as redundancy in both BERT and RoBERTa.

\section{Related Work}

BERT and RoBERTa are large-scale pre-trained language models based on transformers~\cite{vaswani2017attention}.
Despite their groundbreaking power, there have been many papers trying to examine and exploit their over-parameterization.
\citet{michel2019sixteen} and \citet{voita-etal-2019-analyzing} analyze redundancy in attention heads.
Q-BERT~\cite{shen2019q} uses quantization to compress BERT, and LayerDrop~\cite{fan2019reducing} uses group regularization to enable structured pruning at inference time.
On the knowledge distillation side, TinyBERT~\cite{jiao2019tinybert} and DistilBERT~\cite{sanh2019distilbert} both distill BERT into a smaller transformer-based model,
and \citet{tang-etal-2019-natural} distill BERT into even smaller non-transformer-based models.

Our work is inspired by~\citet{cambazoglu2010early}, \citet{teerapittayanon2016branchynet}, and~\citet{huang2018multiscale}, but mainly differs from previous work in that we focus on improving model efficiency with minimal quality degradation.

\begin{table*}[t]
	\centering\resizebox{\textwidth}{!}{
		\begin{tabular}{lc@{\hspace{3pt}}cc@{\hspace{3pt}}cc@{\hspace{3pt}}cc@{\hspace{3pt}}cc@{\hspace{3pt}}cc@{\hspace{5pt}}c}
			\toprule
			& \multicolumn{2}{c}{SST-2} & \multicolumn{2}{c}{MRPC} & \multicolumn{2}{c}{QNLI} & \multicolumn{2}{c}{RTE} & \multicolumn{2}{c}{QQP} & \multicolumn{2}{c}{MNLI-(m/mm)}\\
			\cmidrule(lr){2-3} \cmidrule(lr){4-5} \cmidrule(lr){6-7} \cmidrule(lr){8-9} \cmidrule(lr){10-11} \cmidrule(lr){12-13}
			& Acc & Time & F$_1$ & Time & Acc & Time & Acc & Time & F$_1$ & Time & Acc & Time \\
			\midrule
			\multicolumn{13}{c}{\bf BERT-base} \\
			\midrule
			Baseline 
			& 93.6 & 36.72s & 88.2 & 34.77s & 91.0 & 111.44s & 69.9 & 61.26s & 71.4 & 145min & 83.9/83.0 & 202.84s \\
			DistilBERT & $-$1.4 & $-$40\% & $-$1.1 & $-$40\% & $-$2.6 & $-$40\% & $-$9.4 & $-$40\% & $-$1.1 & $-$40\% & $-$4.5 & $-$40\% \\
			\cmidrule{2-13}
			\multirow{3}{*}{\name}
			& $-$0.2 & $-$21\% & $-$0.3 & $-$14\% & $-$0.1 & $-$15\% & $-$0.4 & $-$9\% & $-$0.0 & $-$24\% & $-$0.0/$-$0.1 & $-$14\%\\
			& $-$0.6 & $-$40\% & $-$1.3 & $-$31\% & $-$0.7 & $-$29\% & $-$0.6 & $-$11\% & $-$0.1 & $-$39\% & $-$0.8/$-$0.7 & $-$25\%\\
			& $-$2.1 & $-$47\% & $-$3.0 & $-$44\% & $-$3.1 & $-$44\% & $-$3.2 & $-$33\% & $-$2.0 & $-$49\% & $-$3.9/$-$3.8 & $-$37\%\\
			\midrule
			\multicolumn{13}{c}{\bf RoBERTa-base} \\
			\midrule
			Baseline
			& 94.3 & 36.73s & 90.4 & 35.24s & 92.4 & 112.96s & 67.5 & 60.14s & 71.8 & 152min & 87.0/86.3 & 198.52s\\
			LayerDrop & $-$1.8 & $-$50\% & - & - & - & - & - & -  & - & - & $-$4.1 & $-$50\% \\
			\cmidrule{2-13}
			\multirow{3}{*}{\name}
			& $+$0.1 & $-$26\% & $+$0.1 & $-$25\% & $-$0.1 & $-$25\% & $-$0.6 & $-$32\%  & $+$0.1 & $-$32\% & $-$0.0/$-$0.0 & $-$19\% \\
			& $-$0.0 & $-$33\% & $+$0.2 & $-$28\% & $-$0.5 & $-$30\% & $-$0.4 & $-$33\%  & $-$0.0 & $-$39\% & $-$0.1/$-$0.3 & $-$23\% \\
			& $-$1.8 & $-$44\% & $-$1.1 & $-$38\% & $-$2.5 & $-$39\% & $-$1.1 & $-$35\%  & $-$0.6 & $-$44\% & $-$3.9/$-$4.1& $-$29\% \\
			\bottomrule
	\end{tabular}}
	\caption{
		Comparison between baseline (original BERT/RoBERTa), \name, and other acceleration methods.
		LayerDrop only reports results on SST-2 and MNLI.
		Time savings of DistilBERT and LayerDrop are estimated by reported model size reduction.
	}
	\label{tab:res}
\end{table*}

\section{Early Exit for BERT inference}

\noindent \name~modifies fine-tuning and inference of BERT models, leaving pre-training unchanged.
It adds one \highway~for each transformer layer.
An inference sample can exit earlier at an \highway, without going through the rest of the transformer layers.
The last \highway~is the classification layer of the original BERT model.

\subsection{\name~at Fine-Tuning}

We start with a pre-trained BERT model with $n$ transformer layers and add $n$ \highways~to it.
For fine-tuning on a downstream task, the loss function of the $i^\text{th}$ \highway~is
\begin{equation}
    L_i(\mcal{D}; \theta) = \frac{1}{|\mcal{D}|} \sum_{(x, y) \in \mcal{D}} H(y, f_i(x; \theta)),
\end{equation}
where $\mcal{D}$ is the fine-tuning training set, $\theta$ is the collection of all parameters, $(x, y)$ is the feature--label pair of a sample, $H$ is the cross-entropy loss function, and $f_i$ is the output of the $i^\text{th}$ \highway.

The network is fine-tuned in two stages:
\begin{enumerate}[leftmargin=*]
	
    \item Update the embedding layer, all transformer layers, and the last \highway~with the loss function $L_n$.
    This stage is identical to BERT fine-tuning in the original paper~\cite{devlin-etal-2019-bert}.
    
    \item Freeze all parameters fine-tuned in the first stage, and then update all but the last \highway~with the loss function $\sum_{i=1}^{n-1}L_i$.
    The reason for freezing parameters of transformer layers is to keep the optimal output quality for the last \highway; otherwise, transformer layers are no longer optimized solely for the last \highway,
    generally worsening its quality.
\end{enumerate}

\subsection{\name~at Inference}

\begin{algorithm}[t]
	\begin{algorithmic}
		\FOR{$i=1$ to $n$}
		\STATE{$z_i = f_i(x; \theta)$}
		\IF{$\mathrm{entropy}(z_i)<S$}
		\RETURN{$z_i$}
		\ENDIF
		\ENDFOR
		\RETURN{$z_n$}
	\end{algorithmic}
	\caption{\name~Inference (Input: $x$)}
	\label{alg}
\end{algorithm}

The way \name~works at inference time is shown in Algorithm \ref{alg}.
We quantify an \highway's confidence in its prediction using the entropy of the output probability distribution $z_i$.
When an input sample $x$ arrives at an \highway, the \highway~compares the entropy of its output distribution $z_i$ with a preset threshold $S$ to determine whether the sample should be returned here or sent to the next transformer layer.

It is clear from both intuition and experimentation that a larger $S$ leads to a faster but less accurate model, and a smaller $S$ leads to a more accurate but slower one.
In our experiments, we choose $S$ based on this principle.

We also explored using ensembles of multiple layers instead of a single layer for the \highway, but this does not bring significant improvements.
The reason is that predictions from different layers are usually highly correlated, and a wrong prediction is unlikely to be ``fixed'' by the other layers.
Therefore, we stick to the simple yet efficient single output layer strategy.

\section{Experiments}

\subsection{Experimental Setup}

We apply \name~to both BERT and RoBERTa, and conduct experiments on six classification datasets from the GLUE benchmark~\cite{wang-etal-2018-glue}:\ SST-2, MRPC, QNLI, RTE, QQP, and MNLI.
Our implementation of \name~is adapted from the HuggingFace Transformers Library~\cite{Wolf2019HuggingFacesTS}.
Inference runtime measurements are performed on a single NVIDIA Tesla P100 graphics card.
Hyperparameters such as hidden-state size, learning rate, fine-tune epoch, and batch size are kept unchanged from the library.
There is no early stopping and the checkpoint after full fine-tuning is chosen.

\subsection{Main Results}
\label{sec:results}

We vary \name's quality--efficiency trade-off by setting different entropy thresholds $S$, and compare the results with other baselines in Table \ref{tab:res}.
Model quality is measured on the test set, and the results are provided by the GLUE evaluation server.
Efficiency is quantified with wall-clock inference runtime\footnote{This includes both CPU and GPU runtime.} on the entire test set, where samples are fed into the model one by one.
For each run of \name~on a dataset, we choose three entropy thresholds $S$ based on quality--efficiency trade-offs on the development set, aiming to demonstrate two cases:\ (1) the maximum runtime savings with minimal performance drop ($<0.5\%$), and (2) the runtime savings with moderate performance drop ($2\%-4\%$).
Chosen $S$ values differ for each dataset.

\begin{figure}[t]
	\centering
	\includegraphics[width=0.49\columnwidth]{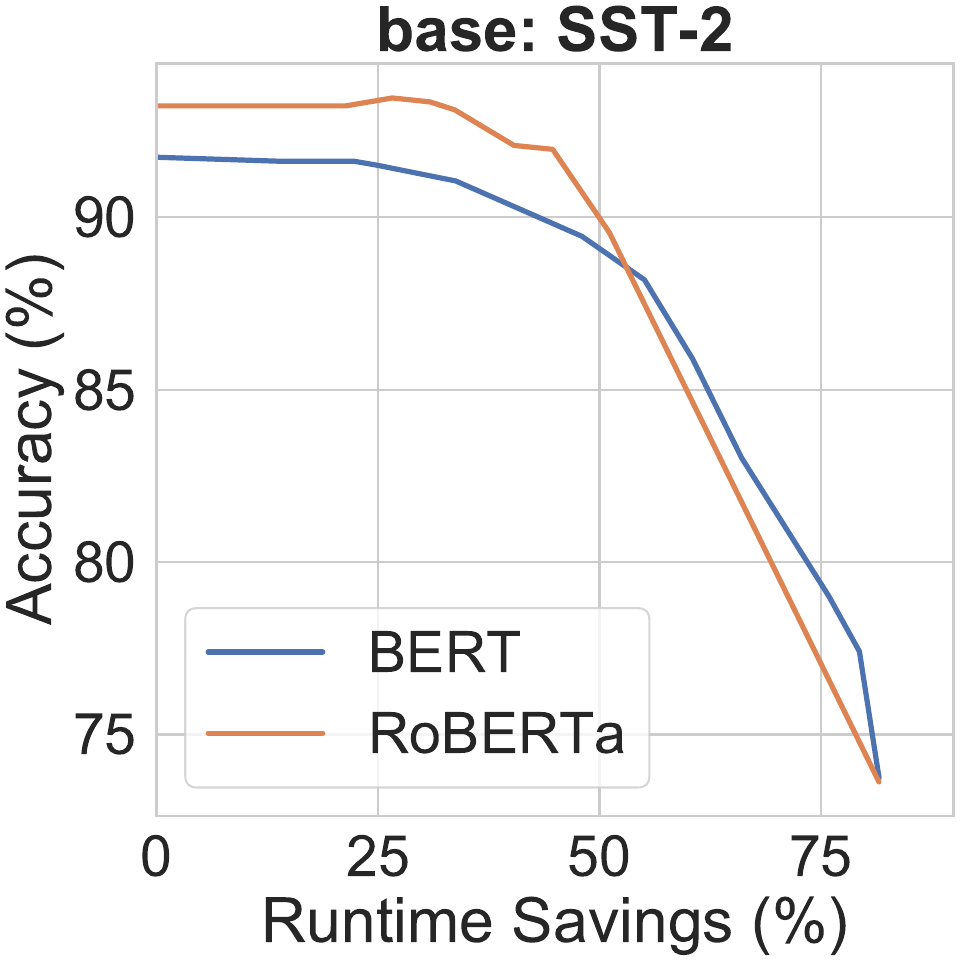}
	\includegraphics[width=0.49\columnwidth]{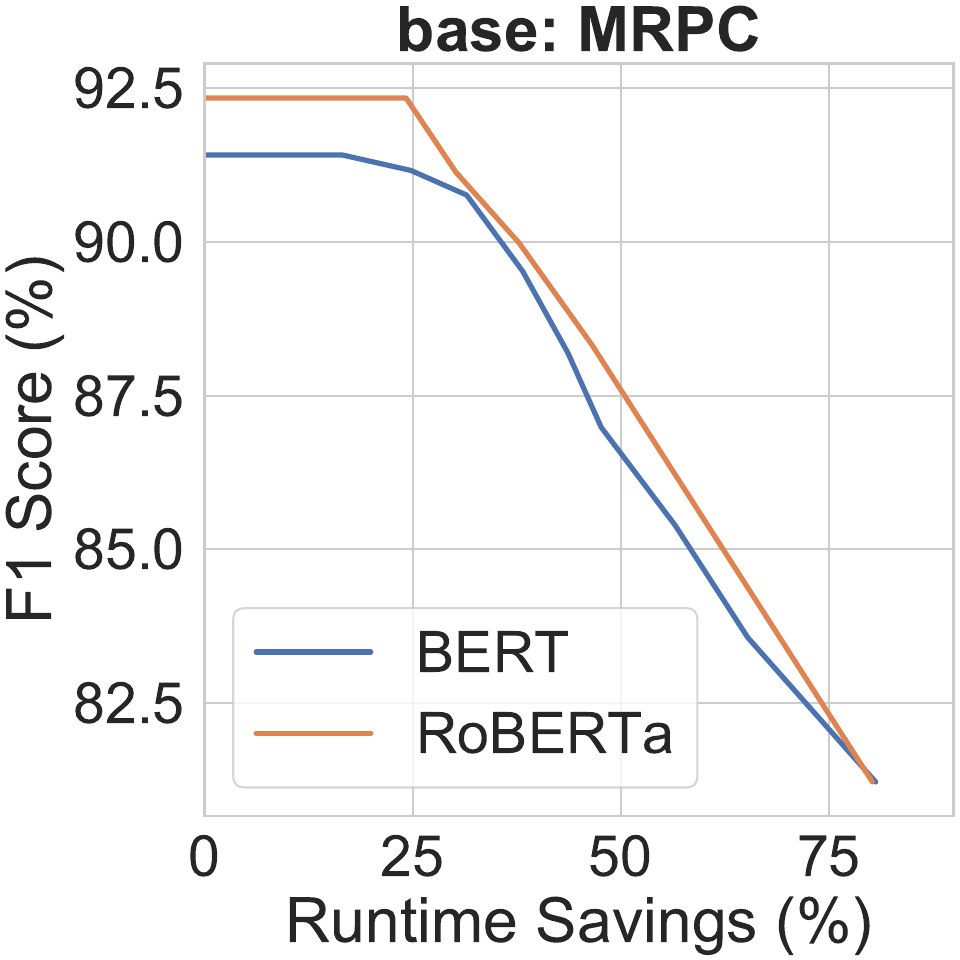}
	\par\medskip
	\includegraphics[width=0.49\columnwidth]{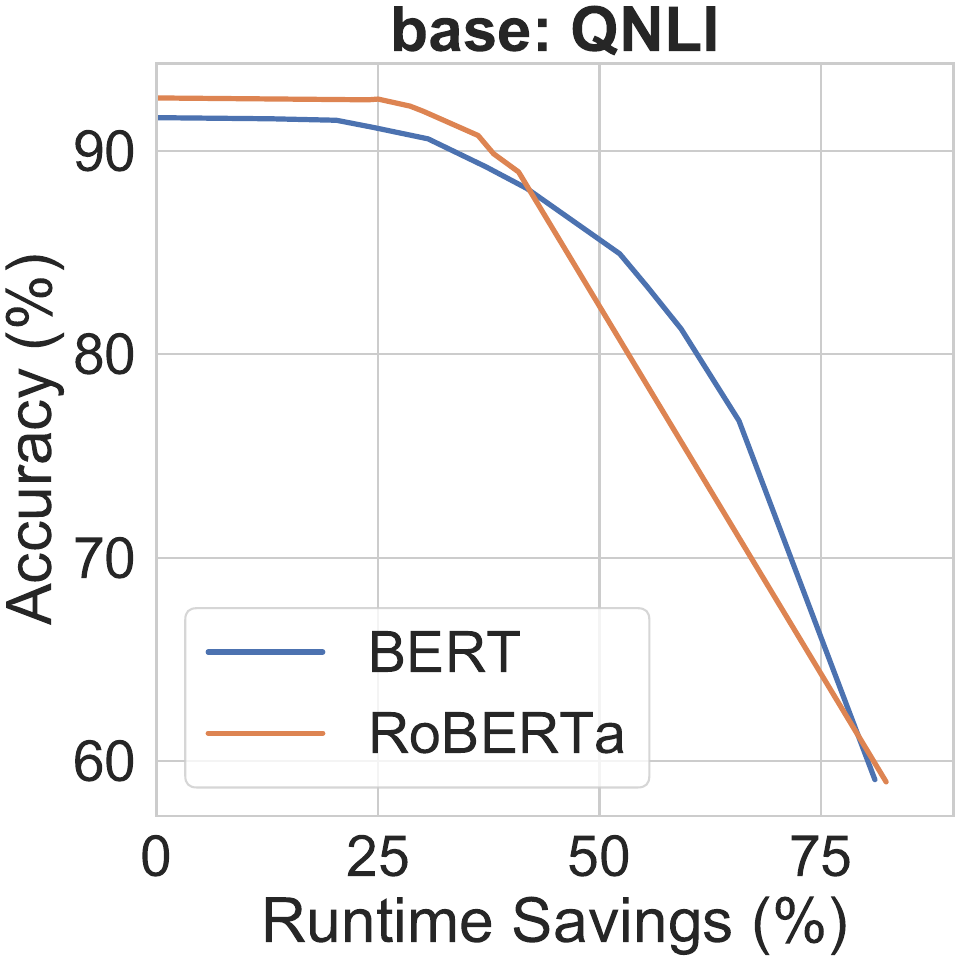}
	\includegraphics[width=0.49\columnwidth]{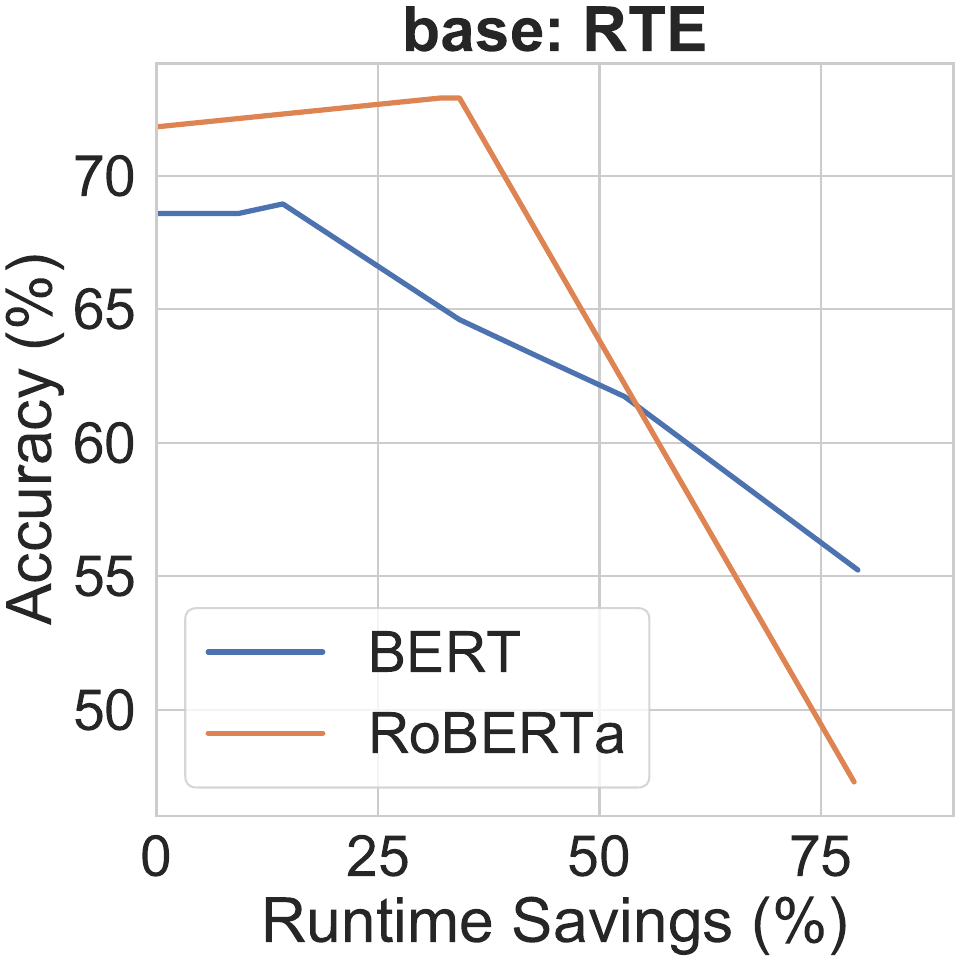}
	\par\medskip
	\includegraphics[width=0.49\columnwidth]{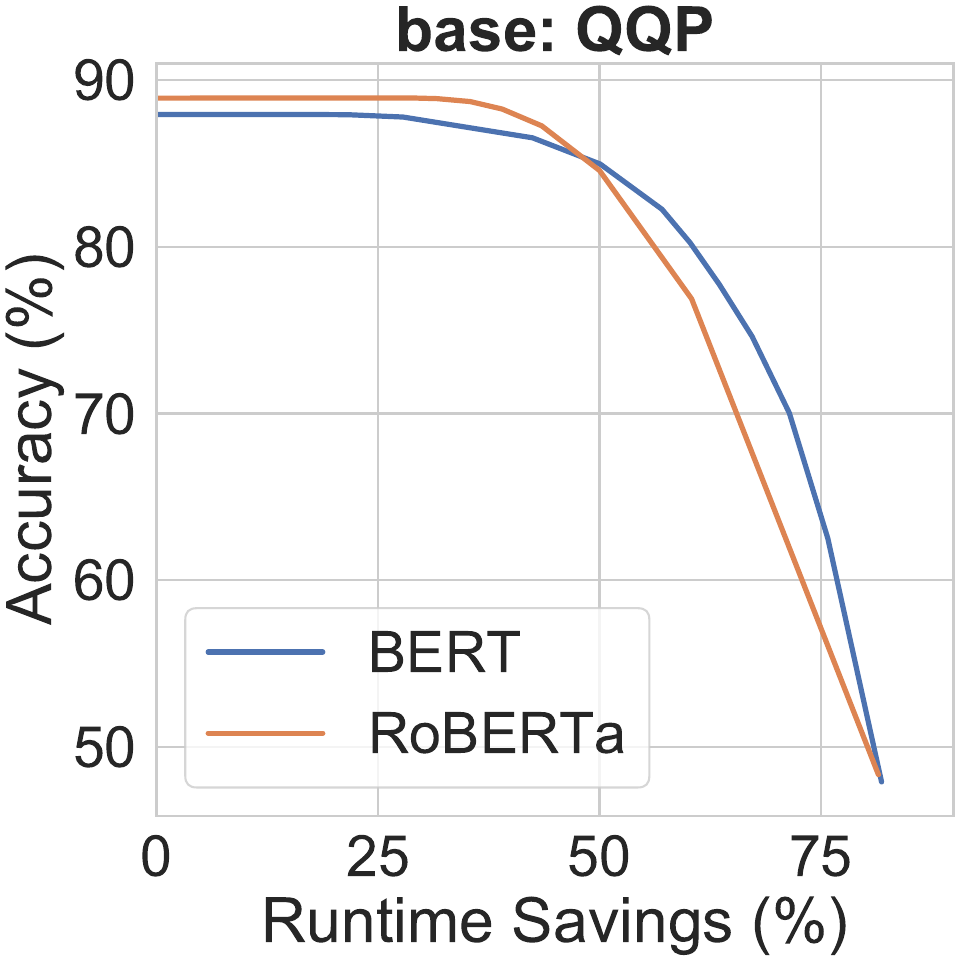}
	\includegraphics[width=0.49\columnwidth]{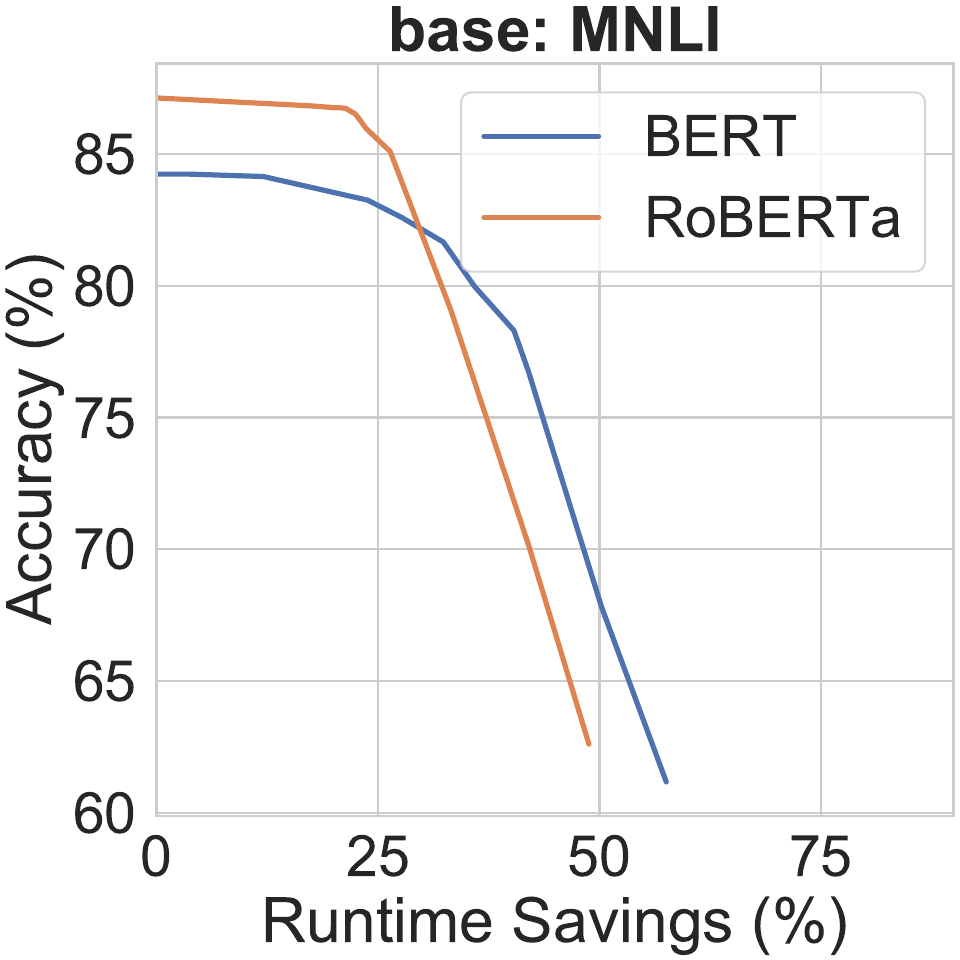}
	\caption{\name~quality and efficiency trade-offs for BERT-base and RoBERTa-base models.}
	\label{fig:tradeoff}
\end{figure}

\begin{figure}[t]
	\centering
	\includegraphics[width=0.49\columnwidth]{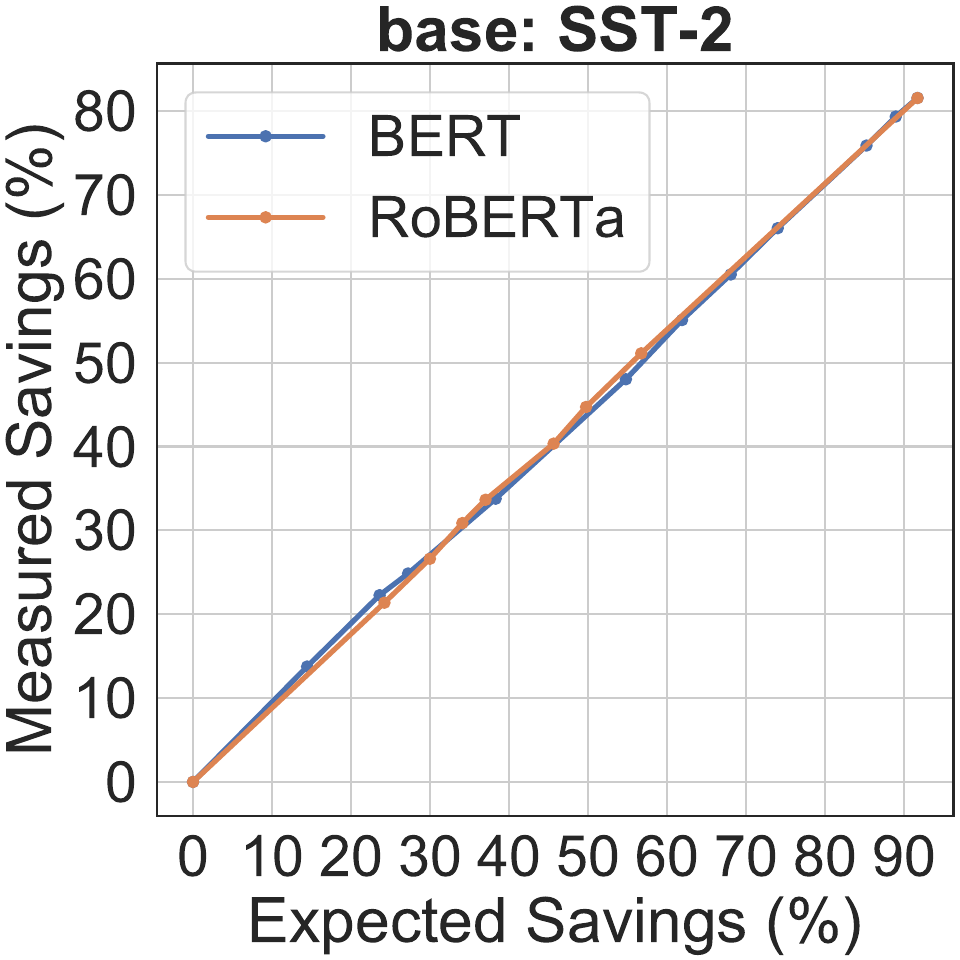}
	\includegraphics[width=0.49\columnwidth]{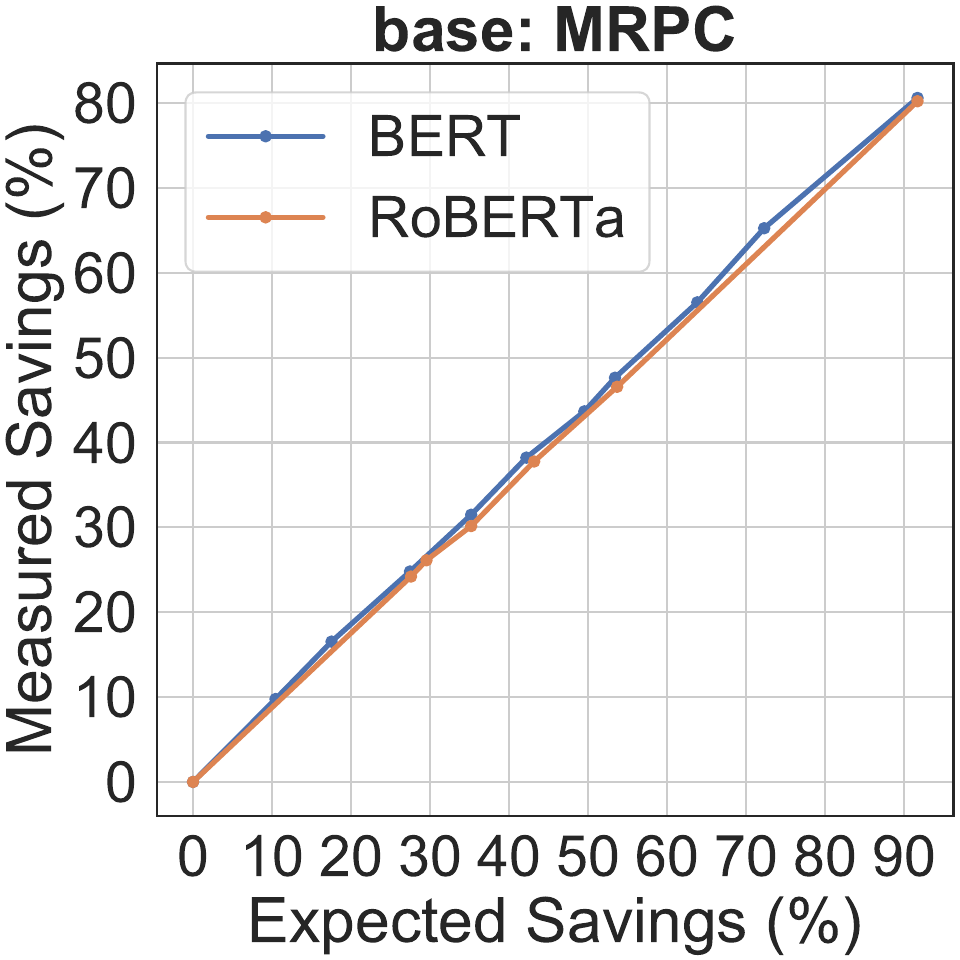}
	\caption{Comparison between expected saving ($x$-axis) and actual measured saving ($y$-axis), using BERT-base and RoBERTa-base models.}
	\label{fig:ers}
\end{figure}

We also visualize the trade-off in Figure \ref{fig:tradeoff}.
Each curve is drawn by interpolating a number of points, each of which corresponds to a different threshold $S$.
Since this only involves a comparison between different settings of \name, runtime is measured on the development set.

From Table \ref{tab:res} and Figure \ref{fig:tradeoff}, we observe the following patterns:

\begin{itemize}[leftmargin=*]
    \item Despite differences in baseline performance, both models show similar patterns on all datasets:\ the performance (accuracy/F$_1$ score) stays (mostly) the same until runtime saving reaches a certain turning point, and then starts to drop gradually. The turning point typically comes earlier for BERT than for RoBERTa, but after the turning point, the performance of RoBERTa drops faster than for BERT. The reason for this will be discussed in Section \ref{sec:layer-wise}.
    
    \item Occasionally, we observe spikes in the curves, e.g., RoBERTa in SST-2, and both BERT and RoBERTa in RTE. We attribute this to possible regularization brought by early exiting and thus smaller effective model sizes, i.e., in some cases, using all transformer layers may not be as good as using only some of them.
\end{itemize}

\noindent Compared with other BERT acceleration methods, \name~has the following two advantages:

\begin{itemize}[leftmargin=*]

	\item Instead of producing a fixed-size smaller model like DistilBERT~\cite{sanh2019distilbert}, \name~produces a series of options for faster inference, which users have the flexibility to choose from, according to their demands.
	
	\item Unlike DistilBERT and LayerDrop~\cite{fan2019reducing}, \name~does not require further pre-training of the transformer model, which is much more time-consuming than fine-tuning.
\end{itemize}

\subsection{Expected Savings}

As the measurement of runtime might not be stable, we propose another metric to capture efficiency, called expected saving, defined as
\begin{equation}
1-  \frac{\sum_{i=1}^{n}i \times N_i}{\sum_{i=1}^{n}n \times N_i},
\end{equation}
where $n$ is the number of layers and $N_i$ is the number of samples exiting at layer $i$.
Intuitively, expected saving is the fraction of transformer layer execution saved by using early exiting.
The advantage of this metric is that it remains invariant between different runs and can be analytically computed.
For validation, we compare this metric with measured saving in Figure \ref{fig:ers}.
Overall, the curves show a linear relationship between expected savings and measured savings, indicating that our reported runtime is a stable measurement of \name's efficiency.

\begin{figure}[t]
	\centering
	\includegraphics[width=0.49\columnwidth]{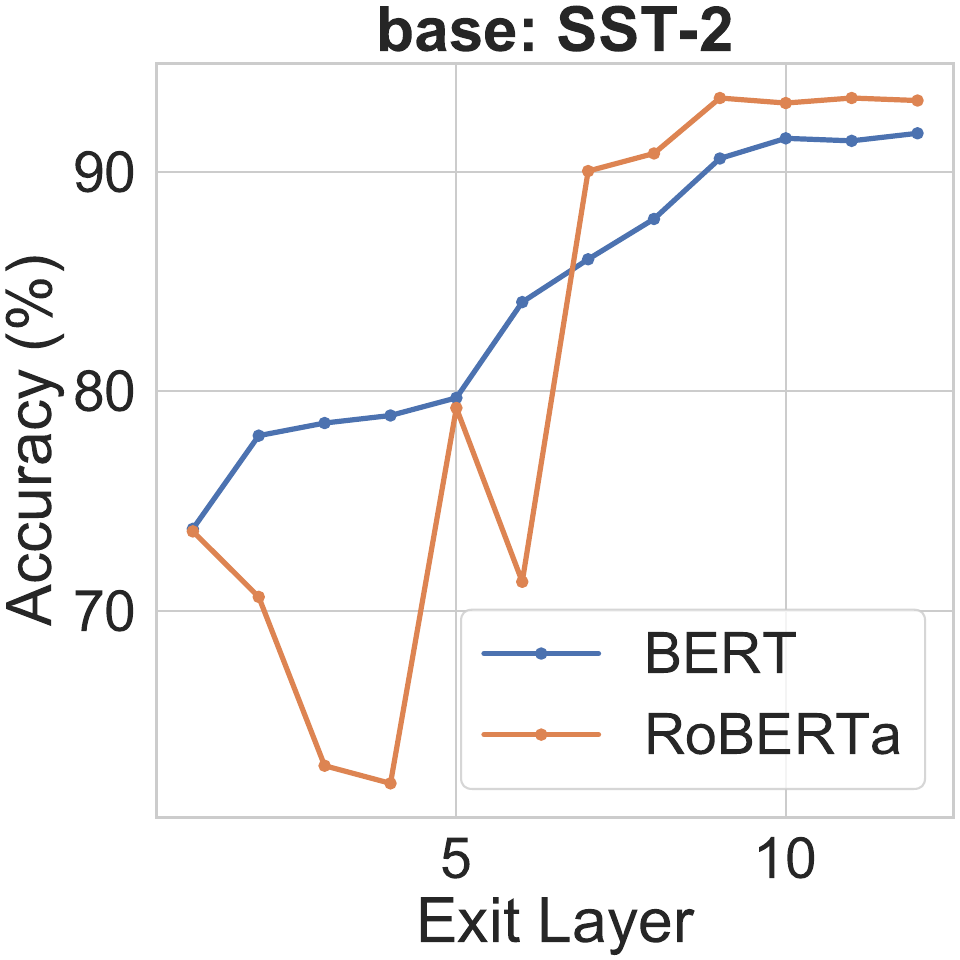}
	\includegraphics[width=0.49\columnwidth]{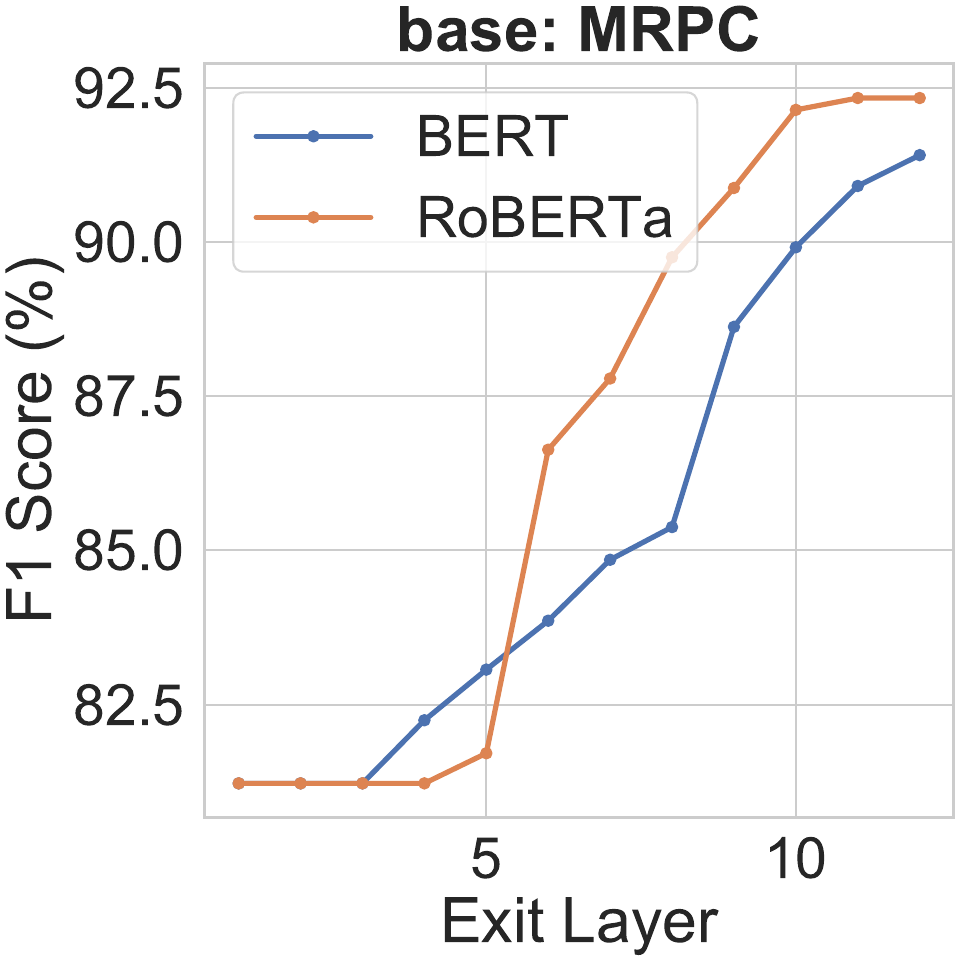}
	\par\medskip
	\includegraphics[width=0.49\columnwidth]{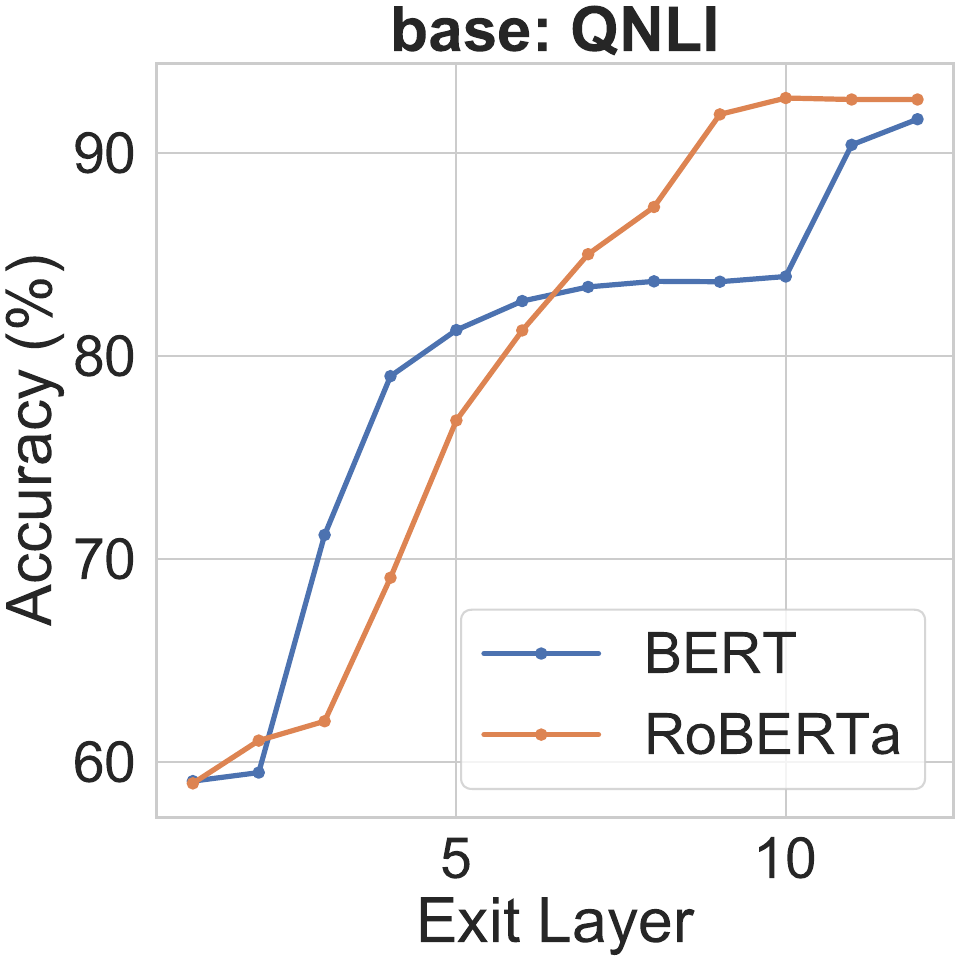}
	\includegraphics[width=0.49\columnwidth]{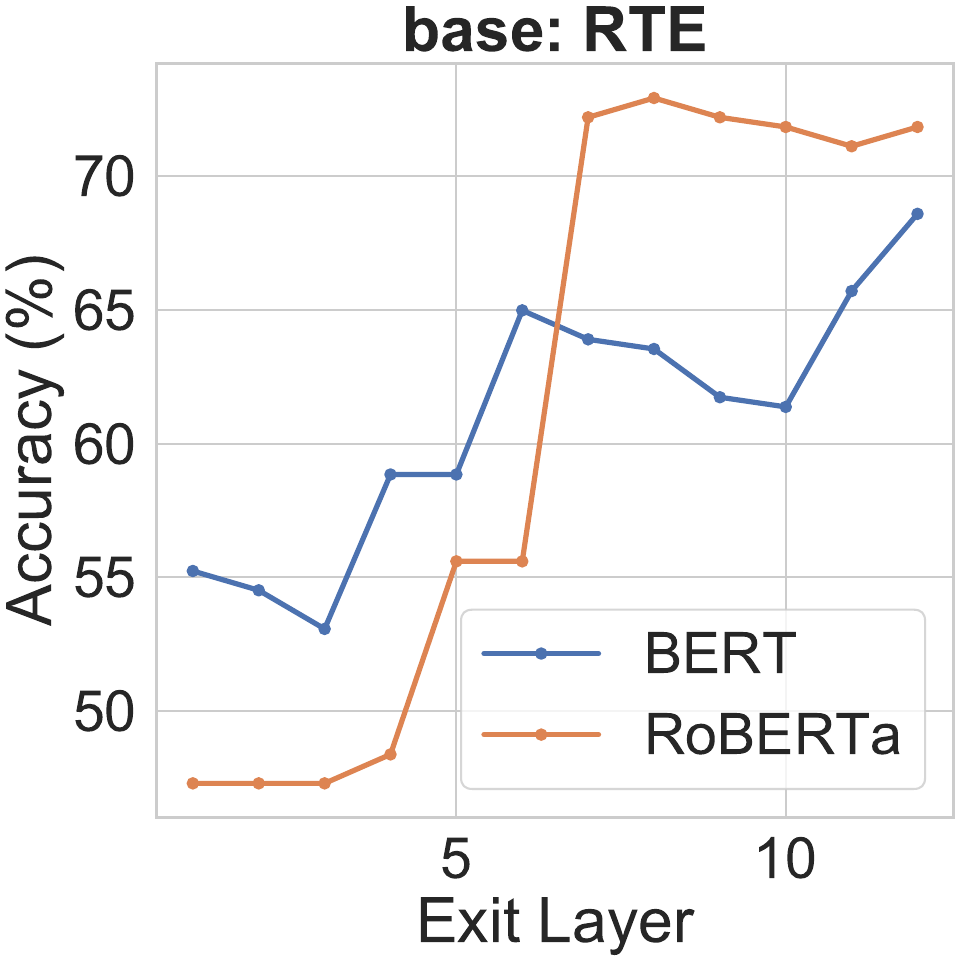}
	\par\medskip
	\includegraphics[width=0.49\columnwidth]{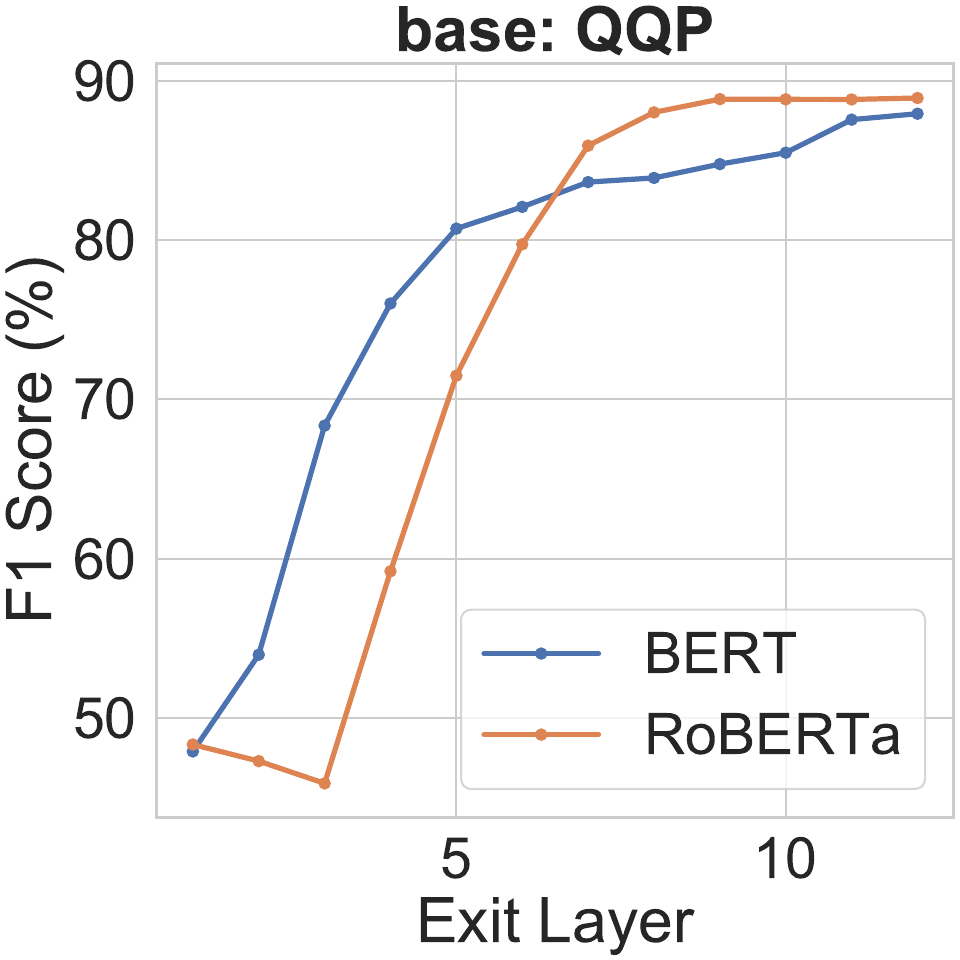}
	\includegraphics[width=0.49\columnwidth]{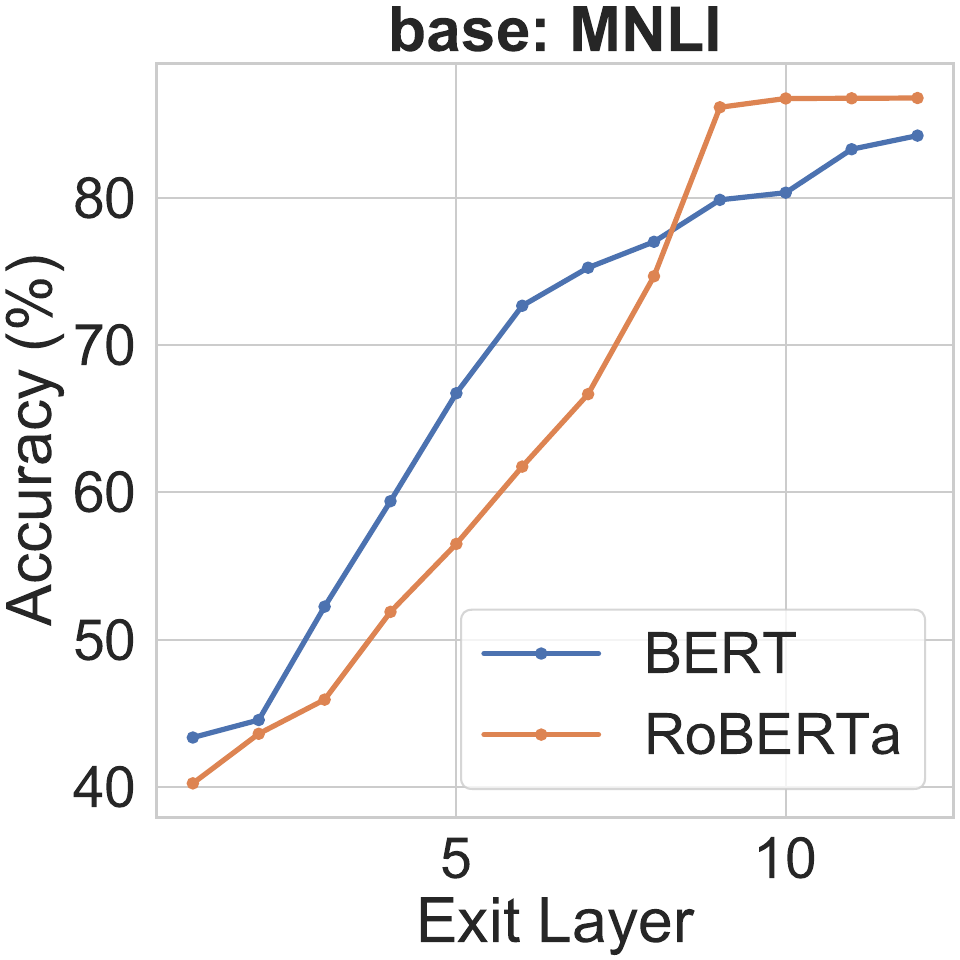}
	\caption{Accuracy of each \highway~for BERT-base and RoBERTa-base.}
	\label{fig:acc-level}
\end{figure}

\begin{figure}[t]
	\centering
	\includegraphics[width=0.49\columnwidth]{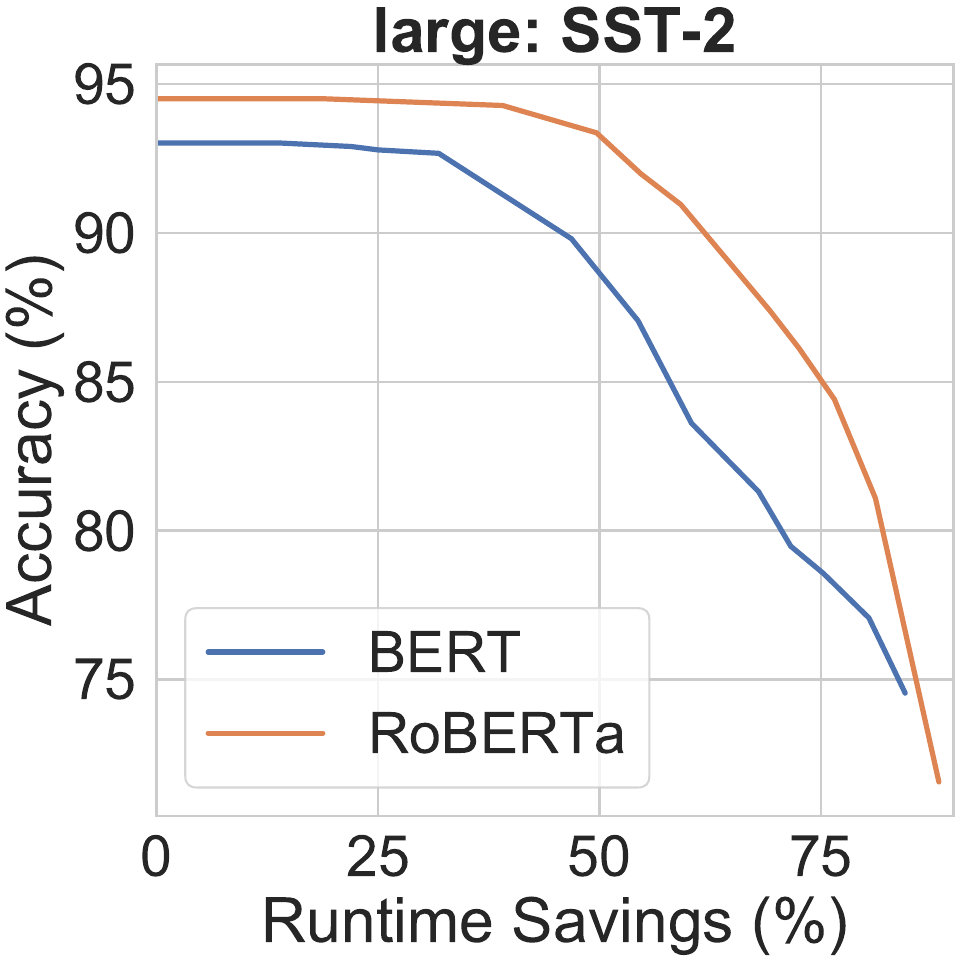}
	\includegraphics[width=0.49\columnwidth]{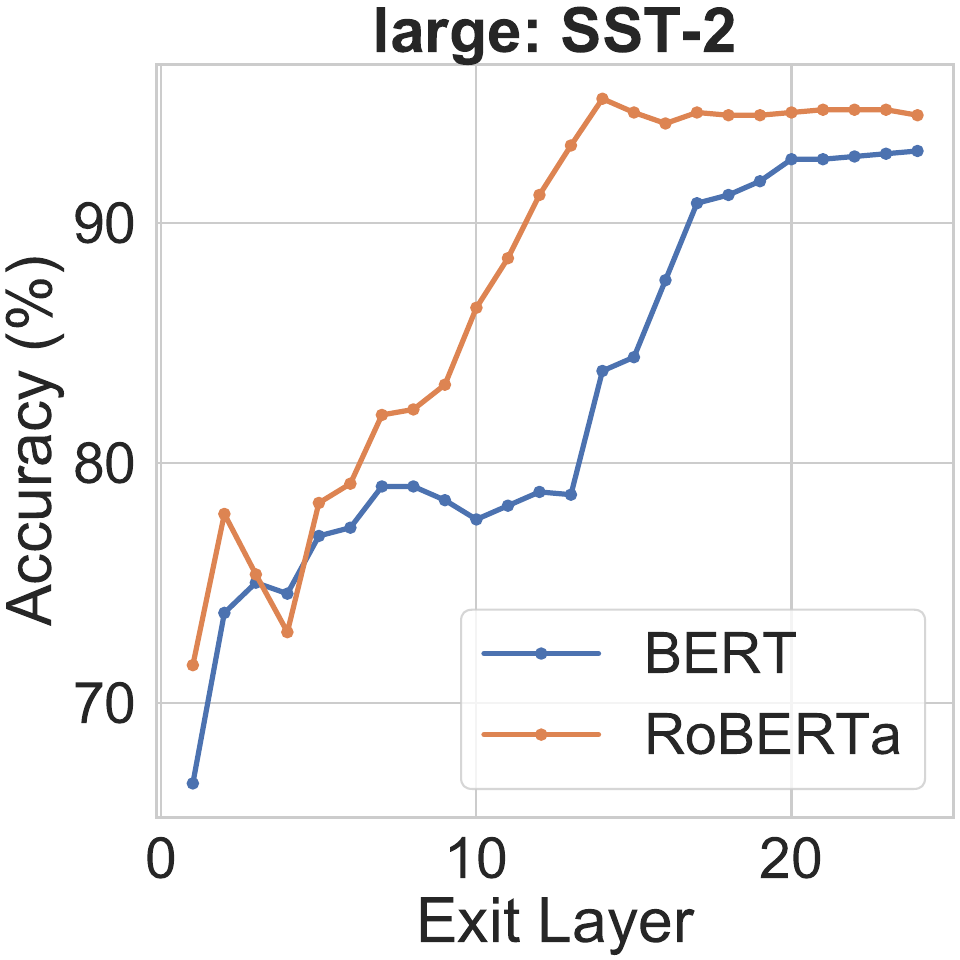}
	\par\medskip
	\includegraphics[width=0.49\columnwidth]{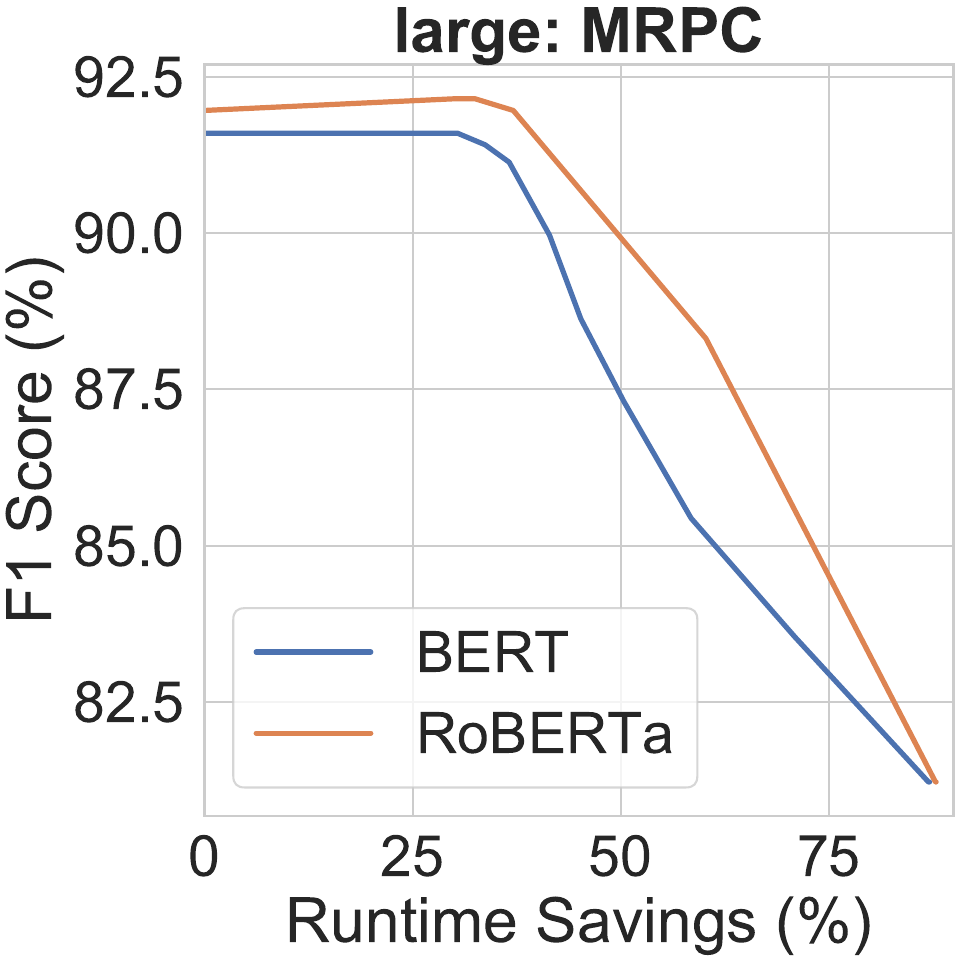}
	\includegraphics[width=0.49\columnwidth]{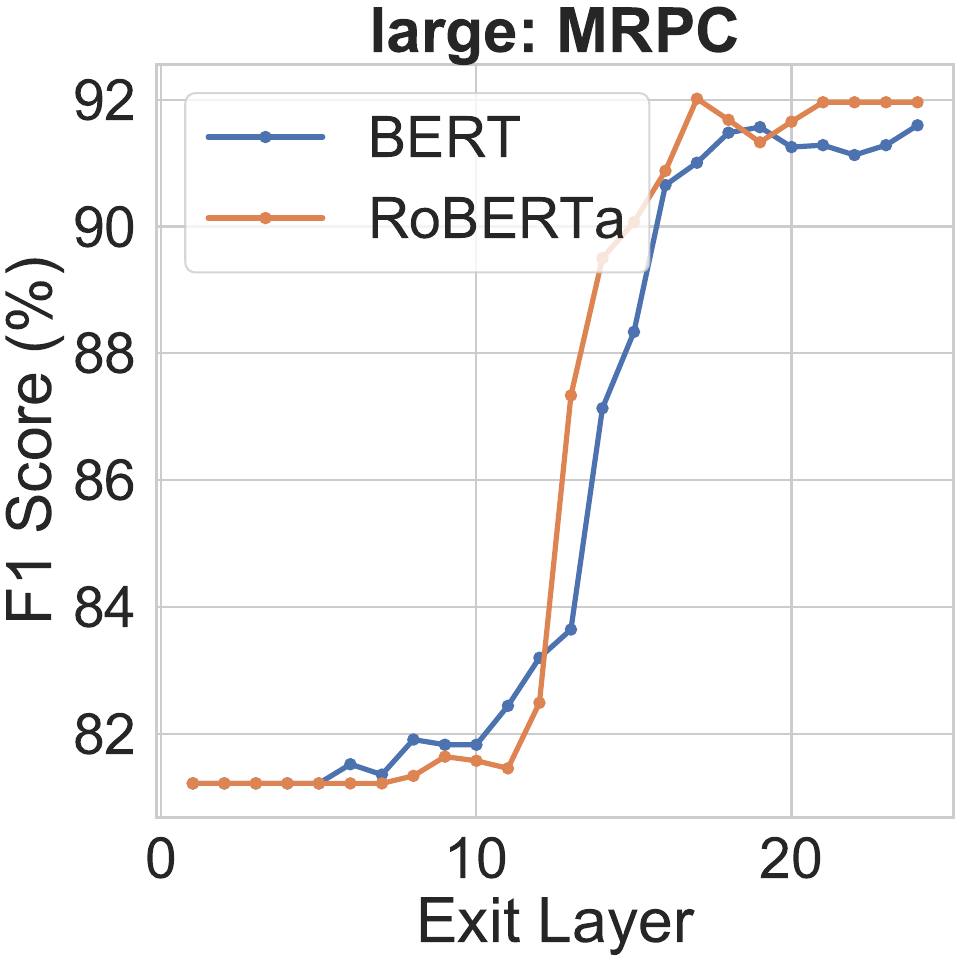}
	\caption{Results for BERT-large and RoBERTa-large.}
	\label{fig:large}
\end{figure}

\subsection{Layerwise Analyses}
\label{sec:layer-wise}

In order to understand the effect of applying \name~to both models, we conduct further analyses on each \highway~layer.
Experiments in this section are also performed on the development set.

\paragraph{Output Performance by Layer.}

For each \highway, we force \textit{all samples} in the development set to exit here, measure the output quality, and visualize the results in Figure \ref{fig:acc-level}.

From the figure, we notice the difference between BERT and RoBERTa.
The output quality of BERT improves at a relatively stable rate as the index of the exit \highway~increases.
The output quality of RoBERTa, on the other hand, stays almost unchanged (or even worsens) for a few layers, then rapidly improves, and reaches a saturation point before BERT does.
This provides an explanation for the phenomenon mentioned in Section \ref{sec:results}:\ on the same dataset, RoBERTa often achieves more runtime savings while maintaining roughly the same output quality, but then quality drops faster after reaching the turning point.

We also show the results for BERT-large and RoBERTa-large in Figure \ref{fig:large}.
From the two plots on the right, we observe signs of redundancy that both BERT-large and RoBERTa-large share:\ the last several layers do not show much improvement compared with the previous layers (performance even drops slightly in some cases).
Such redundancy can also be seen in Figure \ref{fig:acc-level}.

\paragraph{Number of Exiting Samples by Layer.}

\begin{figure}[t]
	\centering
	\hspace{-10pt}
	\includegraphics[width=0.49\columnwidth]{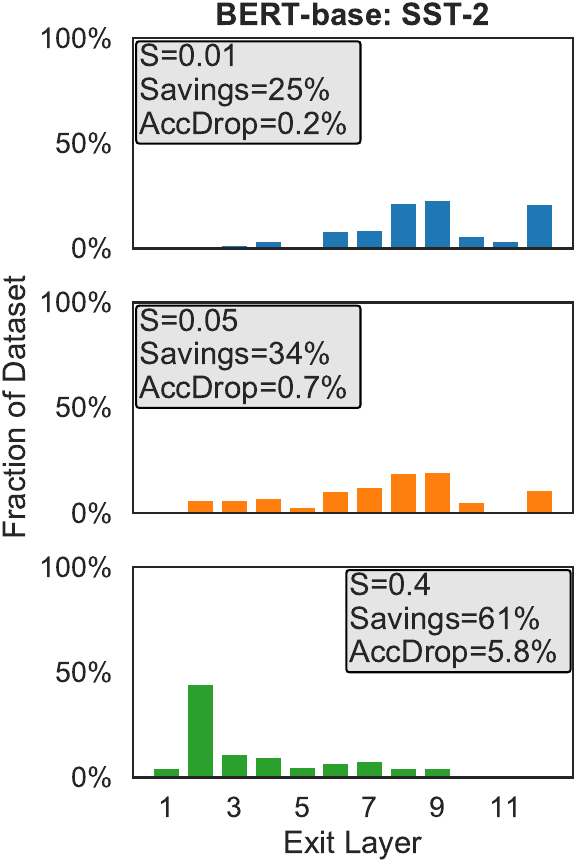}
	\includegraphics[width=0.49\columnwidth]{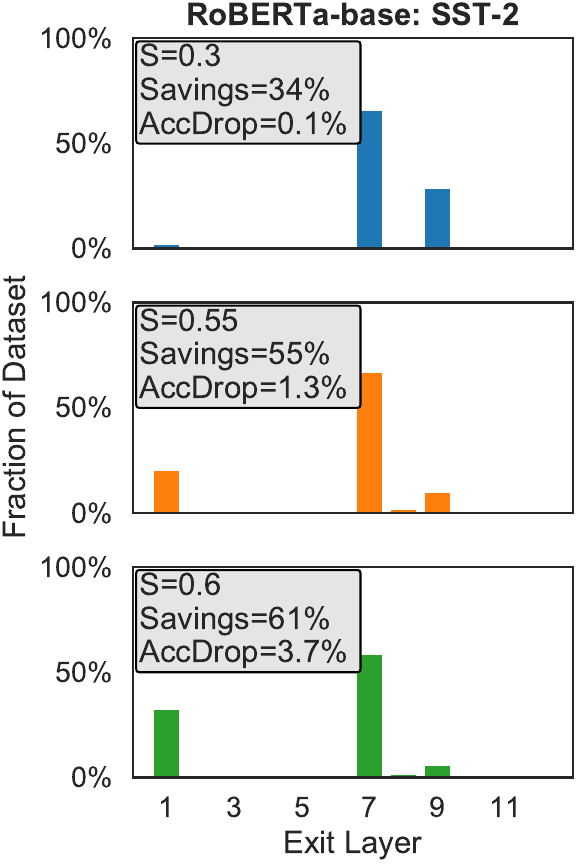}
	\caption{Number of output samples by layer for BERT-base and RoBERTa-base. Each plot represents a separate entropy threshold $S$.}
	\label{fig:sample-layer}
\end{figure}

We further show the fraction of samples exiting at each \highway~for a given entropy threshold in Figure \ref{fig:sample-layer}.

Entropy threshold $S=0$ is the baseline, and all samples exit at the last layer;
as $S$ increases, gradually more samples exit earlier.
Apart from the obvious, we observe additional, interesting patterns:
if a layer does not provide better-quality output than previous layers, such as layer 11 in BERT-base and layers 2--4 and 6 in RoBERTa-base (which can be seen in Figure \ref{fig:acc-level}, top left), it is typically chosen by very few samples;
popular layers are typically those that substantially improve over previous layers, such as layer 7 and 9 in RoBERTa-base.
This shows that an entropy threshold is able to choose the fastest \highway~among those with comparable quality,
and achieves a good trade-off between quality and efficiency.

\section{Conclusions and Future Work}

We propose \name, an effective method that exploits redundancy in BERT models to achieve better quality--efficiency trade-offs.
Experiments demonstrate its ability to accelerate BERT's and RoBERTa's inference by up to $\sim$40\%, and also reveal interesting patterns of different transformer layers in BERT models.

There are a few interesting questions left unanswered in this paper, which would provide interesting future research directions:
(1) DeeBERT's training method, while maintaining good quality in the last \highway, reduces model capacity available for intermediate \highways;
it would be important to look for a method that achieves a better balance between all \highways.
(2) The reasons why some transformer layers appear redundant\footnote{For example, the first and last four layers of RoBERTa-base on SST-2 (Figure \ref{fig:acc-level}, top left).} and why DeeBERT considers some samples easier than others remain unknown;
it would be interesting to further explore relationships between pre-training and layer redundancy, sample complexity and exit layer, and related characteristics.

\section*{Acknowledgment}

We thank anonymous reviewers for their insightful suggestions.
We also gratefully acknowledge funding support from the Natural Sciences and Engineering Research Council (NSERC) of Canada.
Computational resources used in this work were provided, in part, by the Province of Ontario, the Government of Canada through CIFAR, and companies sponsoring the Vector Institute.

\bibliography{acl2020}
\bibliographystyle{acl_natbib}

\end{document}